\documentclass[a4paper,conference]{IEEEtran}
\IEEEoverridecommandlockouts                              %

\title{Scrape, Cut, Paste and Learn: Automated Dataset Generation Applied to Parcel Logistics}

\author{
    Alexander Naumann$^{1,2}$,  %
    Felix Hertlein$^{1}$, %
    Benchun Zhou$^{2}$, %
    Laura D\"orr$^{1,2}$ %
    and 
    Kai Furmans$^{2}$%
    \thanks{
        $^{1}$The authors are with the 
        FZI Research Center for Information Technology, Karlsruhe, Germany
            {\tt\small \{anaumann, hertlein, doerr\}@fzi.de}\newline%
        \indent $^{2}$The authors are with the Institute for Material Handling and Logistics, 
        Karlsruhe Institute of Technology (KIT),
        Karlsruhe, Germany
            {\tt\small \{benchun.zhou, kai.furmans\}@kit.edu}%
    }
}

\usepackage{booktabs}
\usepackage{wrapfig}

\usepackage[all]{nowidow} %
\usepackage{amsmath}
\usepackage{amsfonts}
\usepackage{amssymb}
\usepackage[pdftex]{graphicx}
\graphicspath{{graphics/}}
\usepackage{trfsigns}
\usepackage{caption} %
\usepackage{subcaption} %
\usepackage{enumitem}
\usepackage{hyperref}
\usepackage{amsthm}
\usepackage{xcolor}
\usepackage{calc}

\usepackage[acronym]{glossaries}
\DeclareCaptionLabelFormat{blank}{}
\usepackage{siunitx}
\usepackage{tikz}
\usetikzlibrary{patterns}
\usetikzlibrary{calc,shapes.arrows}
\usetikzlibrary{arrows.meta}
\usetikzlibrary{hobby}
\usepackage{pgfplots}
\usepackage{placeins}
\usepackage[normalem]{ulem}  %
\usepackage{flushend}   %

\usepackage[framemethod=tikz]{mdframed}

\usepackage{xstring}
\usepackage{catchfile}

\usepackage[style=ieee,  %
            doi=false,
            url=false,
            isbn=false,
            mincitenames=1,
            maxcitenames=1,
            minbibnames=6,
            maxbibnames=6,
            backend=biber]{biblatex}  %

\DeclareSourcemap{
  \maps[datatype=bibtex, overwrite]{
    \map{
      \step[fieldset=address, null]
      \step[fieldset=location, null]
      \step[fieldset=note, null]
      \step[fieldset=series, null]
      \step[fieldset=issn, null]
      \step[fieldset=pages, null]
      \step[fieldset=editor, null]
      \step[fieldset=primaryclass, null] 
      \step[fieldset=publisher, null]
      \step[fieldset=volume, null]
      \step[fieldset=issue, null]
      \step[fieldset=number, null]
      \step[fieldset=eventtitle, null]
    }
    \map[overwrite]{
      \pertype{unpublished} %
       \step[fieldsource=eprint,]
       \step[fieldset=note, fieldvalue={arXiv: }, append]
       \step[fieldset=note, origfieldval, append]
    }
    \map{
      \step[fieldsource=booktitle, final] %
      \step[fieldset=archiveprefix, null] %
      \step[fieldset=arxivid, null] %
      \step[fieldset=eprint, null] %
    }
    \map{
      \step[fieldsource=journaltitle, final] %
      \step[fieldset=archiveprefix, null] %
      \step[fieldset=arxivid, null] %
      \step[fieldset=eprint, null] %
    }
    \map[overwrite=true]{ %
       \pertype{online}  %
       \step[fieldset=urldate,fieldvalue={2022-07-08}]
    } 
  }
}

\newcommand{\citet}[1]{\citetitle{#1}}  %

\renewcommand{\refsection}[1]{Sec.~\ref{#1}}

\newcommand{\reffigure}[1]{Fig.~\ref{#1}}
\newcommand{\reftable}[1]{Table~\ref{#1}}

\newcommand{\idest}{i.e.\ } %

\newglossaryentry{i4.0}{name={Industry 4.0}, description={Industry 4.0}}
\newglossaryentry{forklift}{name={forklift truck}, description={}, plural={forklift trucks}}
\newacronym[longplural={automated guided vehicles}]{agv}{AGV}{automated guided vehicle}

\newacronym[longplural={time-of-flight cameras}]{tofc}{ToF camera}{time-of-flight camera}
\newglossaryentry{kinect}{name={MS Kinect}, description={}}
\newacronym{pmd}{PMD}{Photonic Mixing Device}
\newacronym{roi}{ROI}{Region of Interest}
\newacronym{iou}{IoU}{Intersection over Union}
\newacronym{ap}{AP}{Average Precision}
\newglossaryentry{ar}{name={Augmented Reality}, description={}}
\newglossaryentry{rgbd}{name={RGB-D}, description={}}
\newglossaryentry{opencv}{name={OpenCV}, description={}}
\newacronym[longplural={Light Detection And Ranging}]{lidar}{LiDAR}{Light Detection And Ranging}

\newacronym{ransac}{RANSAC}{Random Sampling Consensus}
\newglossaryentry{sota}{name={state-of-the-art}, description={}}
\newacronym{nn}{NN}{Neural Network}
\newacronym{cnn}{CNN}{Convolutional Neural Network}
\newacronym{gcn}{GCN}{Graph Convolutional Neural Network}
\newacronym{zgcn}{0N-GCN}{Zero-Neighbor Graph Convolutional Network}
\newacronym{dbscan}{DBSCAN}{Density-Based Spatial Clustering of Applications with Noise}
\newacronym{sgdm}{SGD+M}{Stochastic Gradient Descent with Momentum}

\newacronym[longplural={Frames per Second}]{fps}{FPS}{Frame per Second}
\newacronym{eu}{EU}{European Union}
\newglossaryentry{poc}{name={proof of concept}, description={}}
\newacronym{gso}{GSO}{Google Scanned Objects}
\newacronym{lld}{LLD}{Large Logo Dataset}

\newcommand{\maskrcnn}{Mask {R-CNN}}
\newcommand{\backbone}{ResNet-50-FPN}
\newcommand{\openimages}{OpenImages}
\newcommand{\parcelds}{Parcel2D}
\newcommand{\dsplain}{Parcel2D Plain}
\newcommand{\dsmanual}{Parcel2D Manual}
\newcommand{\dsmask}{Parcel2D CNN}
\newcommand{\dsreal}{Parcel2D Real}

\newcommand{\shapenetsem}{ShapeNetSem}

\newcommand{\apfifty}{$\text{AP}_{50}$}
\newcommand{\apseventy}{$\text{AP}_{75}$}
\newcommand{\codeurl}{\href{https://a-nau.github.io/parcel2d}{https://a-nau.github.io/parcel2d}}
\usepackage{etoolbox}

\begin{document}
\maketitle
\thispagestyle{empty}
\pagestyle{empty}
\begingroup
\let\clearpage\relax  %
\begin{abstract}
    State-of-the-art approaches in computer vision heavily rely on sufficiently large training datasets.
    For real-world applications, obtaining such a dataset is usually a tedious task.
    In this paper, we present a fully automated pipeline to generate a synthetic dataset for instance segmentation in four steps.
    In contrast to existing work, our pipeline covers every step from data acquisition to the final dataset.
    We first scrape images for the objects of interest from popular image search engines and
    since we rely only on text-based queries the resulting data comprises a wide variety of images.
    Hence, image selection is necessary as a second step.
    This approach of image scraping and selection relaxes the need for a real-world domain-specific dataset that must be either publicly available or created for this purpose.
    We employ an object-agnostic background removal model and compare three different methods for image selection: Object-agnostic pre-processing, manual image selection and \acrshort{cnn}-based image selection.
    In the third step, we generate random arrangements of the object of interest and distractors on arbitrary backgrounds. %
    Finally, the composition of the images is done by pasting the objects using four different blending methods.
    We present a case study for our dataset generation approach by considering parcel segmentation.
    For the evaluation we created a dataset of parcel photos that were annotated automatically.
    We find that
    (1) our dataset generation pipeline allows a successful transfer to real test images (Mask AP 86.2), 
    (2) a very accurate image selection process - in contrast to human intuition - is not crucial and a broader category definition can help to bridge the domain gap,
    (3) the usage of blending methods is beneficial compared to simple copy-and-paste.
    We made our full code for scraping, image composition and training publicly available at \codeurl{}%
    .
\end{abstract}

\section{Introduction}

Common computer vision tasks, such as instance detection or segmentation have a tremendous potential to help the automation of processes in many industries.
For instance, those techniques can be applied for process monitoring or quality control \cite{nocetiMulticameraSystemDamage2018}.
However, since the object of interest can vary widely depending on the underlying use-case, the availability of a ready-to-use dataset of sufficient size is a common problem in practice.
Manual data acquisition and annotation is a time-consuming and costly task, which is why synthetic datasets have become more and more popular \cite{dwibediCutPasteLearn2017}.
When training on a synthetic dataset, with the goal of employing the trained \gls{cnn} for the real use-case application, the domain gap between the synthetic and the real images has to be taken into account \cite{mensinkFactorsInfluenceTransfer2021}.
A promising approach for quick and efficient synthetic dataset generation was presented by \textcite{dwibediCutPasteLearn2017}.
They randomly paste objects and distractors onto background images, while using different blending methods to reduce the influence of local pasting artifacts.
\begin{figure}[!t]
        \centering
        \includegraphics[width=0.90\linewidth]{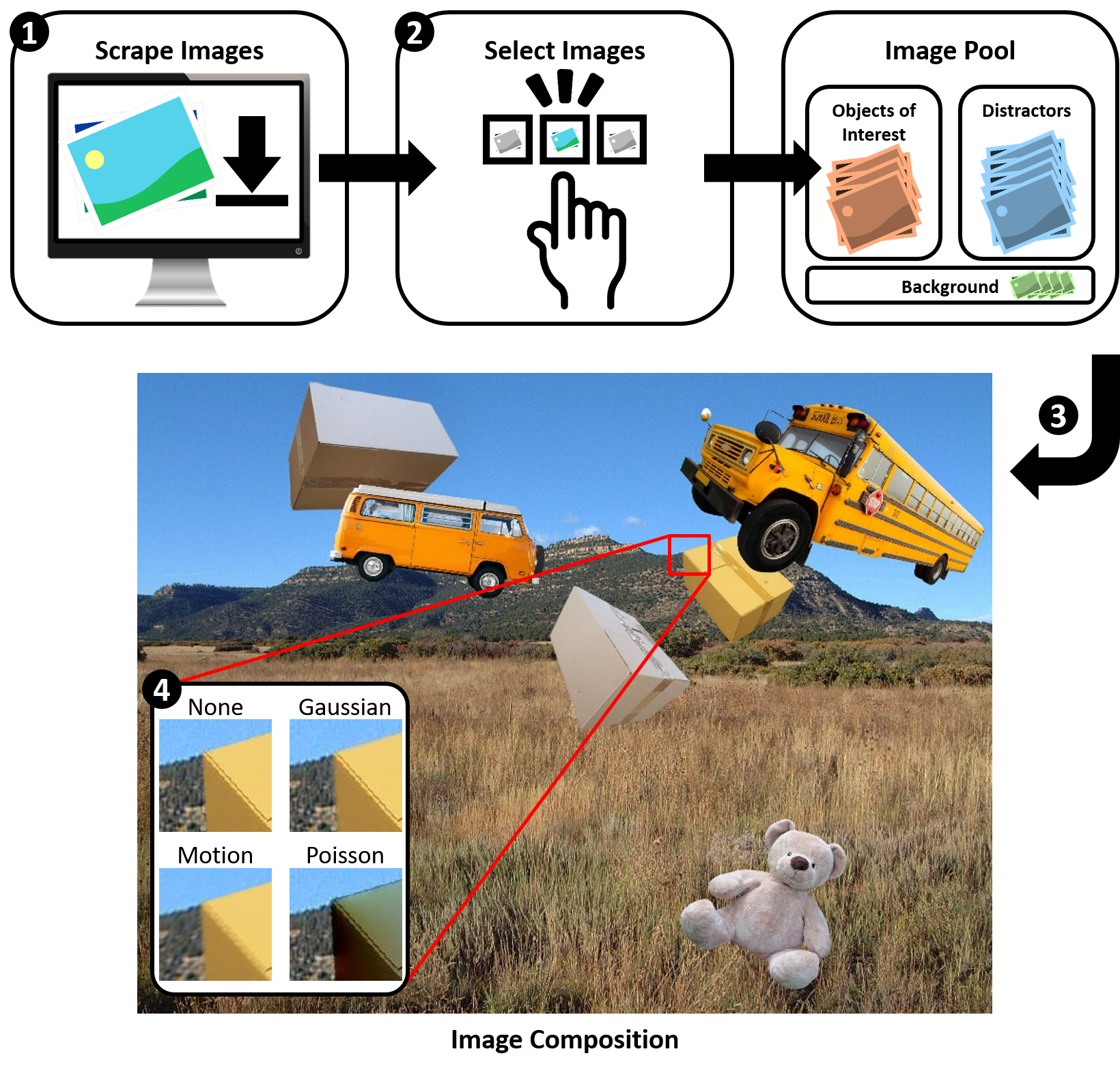}
        \caption{
            Overview of our dataset generation pipeline:
            (1) We scrape images from popular image search engines. 
            (2) We use and compare three different methods for image selection, \idest basic pre-processing, manual selection and CNN-based selection. 
            (3) The objects of interest and the distractors are pasted onto a randomly selected background.
            (4) We use four different blending methods to ensure invariance to local pasting artifacts as suggested by \textcite{dwibediCutPasteLearn2017}.
            }
        \label{fig:overview}
\end{figure}
We extend this approach by adding an automated image selection pipeline as visualized in \reffigure{fig:overview}.
Thus, our pipeline is fully automated and the creation of synthetic datasets is facilitated further.
Note, that not only dataset creation, but also a dataset update, e.g. after a domain shift, is feasible.

We present our results on a case study of parcel detection.
Parcel detection and segmentation is highly relevant in industry since it can help to automate and monitor supply chains \cite{clausenParcelTrackingDetection2019}.
A smoothly running supply chain is crucial for manufacturing industries, pharmaceutical companies and far beyond. 
Furthermore, having a pre-trained backbone that is a strong feature extractor for the application use-case facilitates the development of solutions for downstream tasks, such as keypoint detection or 3D reconstruction.
Our main contributions are:
\begin{itemize}
    \item we extend \cite{dwibediCutPasteLearn2017} by adding image scraping and different image selection methods,
    \item we analyze the influence of the image selection method on the capacity for transfer learning,
    \item we present a real-world dataset of parcel images that is used for evaluation, and
    \item we make our code publicly available, to facilitate the generation of tailored datasets for custom domains.
\end{itemize}

The paper is organized as follows.
We present related literature in \refsection{sec:related_work}.
Subsequently, we describe our dataset generation approach in \refsection{sec:datagen}.
The evaluation is presented in \refsection{sec:eval} and 
the paper concludes with \refsection{sec:conclusion}.

\section{Related Work}
\label{sec:related_work}

The idea of generating an artificial training dataset is widespread, due to the high cost that incur for capturing and annotating a tailor-made dataset for a use-case. 
We first present relevant literature regarding the creation of artificial datasets and subsequently delve into the application area of logistics.

~\\
\vspace{-5mm}

\paragraph*{Artificial Dataset Generation}

Artificial datasets can either be rendered or composed.
When rendering images, we can carefully choose a desired image layout and easily generate a multiplicity of annotations - even the ones that are very costly to obtain, such as 3D annotations.
BlenderProc \cite{denningerBlenderProc2019} is a procedural Blender\footnote{See \href{https://www.blender.org/}{https://www.blender.org/}.} pipeline that enables photorealistic renderings to create synthetic datasets.
Examples for popular rendered datasets include \cite{songSemanticSceneCompletion2017, zhengStructured3DLargePhotoRealistic2020}.

In contrast to that, image datasets can also be generated by composition.
Image composition is the task of seamlessly combining two images by cutting a foreground object from one image and pasting it onto another image.
This is an important task in computer vision with a wide range of applications.
\textcite{niuMakingImagesReal2021} present a comprehensive survey on the topic, and we refer to them for details on applications and subtasks included in image composition.
For our work, we focus on simple image composition and neglect effects that might make images look unrealistic to humans, as this has proven to be sufficient for training the backbone of a neural network \cite{ghiasiSimpleCopyPasteStrong2021}.
More explicitly, inconsistencies introduced by incompatible colors, unreasonable illumination, mismatching size of objects, or their location are not considered.

\textcite{dwibediCutPasteLearn2017} present a procedure to generate a targeted dataset for instance segmentation.
As input, a set of images for each category, picturing solely the object of interest with a modest background, is needed.
They recommend diverse viewpoints, in order to enable detection from diverse viewpoints as well.
A foreground background segmentation network is trained to obtain segmentation masks for the foreground objects.
In addition, suitable background images need to be chosen.
Afterwards, objects are cut out with their mask from the images and pasted onto a background image.
\citeauthor{dwibediCutPasteLearn2017} ensure invariance to local artifacts from pasting by applying a set of blending methods.
The exact same images are synthesized multiple times, where only the blending method varies.
They show that this method enables training a neural network for instance segmentation and that combining the synthetic data with only \SI{10}{\%} of the real training data surpasses the performance compared to training on all real data.
\textcite{ghiasiSimpleCopyPasteStrong2021} present a similar technique, however, they use existing annotated datasets as their source for both the foreground and the background and found scale jittering to be very efficient.
First two images within a dataset are randomly chosen and their scale is jittered.
Subsequently, objects from one image are cut out by using their given annotated mask and pasted randomly onto the second image.
During this process annotations within the second image are adjusted accordingly, \idest adjusted for occlusion.
They do not use geometric transformations such as rotation and find Gaussian blurring not to be beneficial.
\citeauthor{ghiasiSimpleCopyPasteStrong2021} conclude that their method is highly effective and robust.
\textcite{mensinkFactorsInfluenceTransfer2021} present a study on the influence of several factors on the performance for transfer learning.
They find that the image domain is the most important factor and that the target dataset should be contained in the source dataset to achieve best results.

In our work, we follow an approach similar to \textcite{dwibediCutPasteLearn2017}, however, fully automate the foreground object image retrieval by using web scraping and a pre-processing pipeline.

~\\
\vspace{-5mm}

\paragraph*{Applications in Logistics.}
Work on the plane-wise segmentation of parcels, without the need for a custom training dataset was presented by \textcite{naumannRefinedPlaneSegmentation2020}.
Plane segmentation information is combined with contour detection to generate plane-level segmentations.
Small load carriers have been targeted using synthetic training data \cite{mayershoferFullySyntheticTrainingIndustrial2020}.
Furthermore, the problem of packaging structure recognition has been tackled \cite{hinxlageLadungstragerzahlungSmartphone2018, dorrFullyAutomatedPackagingStructure2020, dorrTetraPackNetFourCornerBasedObject2021}.
Packaging structure recognition aims at localizing and counting small load carriers that are stacked onto a pallet.

\section{Dataset Generation}
\label{sec:datagen}

Our dataset generation approach is based on \textcite{dwibediCutPasteLearn2017}.
We follow a similar procedure, apart from the data acquisition approach.
This section is organized as follows:
In \refsection{sec:datagen:scraping}, we explain the data acquisition through web scraping.
Subsequently, we present three different image selection methods which yield three different datasets in \refsection{sec:data:selection}.
The image generation is explained in \refsection{sec:data:generation} and finally we present our real dataset in \refsection{sec:data:real}.

\subsection{Image Scraping}
\label{sec:datagen:scraping}

In order to generate a synthetic dataset, it is crucial to have a sufficiently large set of images picturing the object of interest.
We approach this problem by scraping images from popular image search engines.
We use four different search engines:
\begin{itemize}
    \item Google Images: \href{https://images.google.com/}{images.google.com},
    \item Bing Images: \href{https://www.bing.com/images}{bing.com/images},
    \item Yahoo Images: \href{https://images.search.yahoo.com/}{images.search.yahoo.com} and 
    \item Baidu Images: \href{https://image.baidu.com/}{image.baidu.com}.
\end{itemize}
We scraped images for the object of interest, \idest parcels, and for distractor objects.
The full source code, including a Dockerized web application is available at \codeurl{}.
\paragraph*{Objects of interest}
For the objects of interest, \idest the parcels, we chose 9 different search queries that all represent the same object category: \textit{parcel,
parcel package,
parcel amazon,
packet post,
packing carton,
packing box,
carton box,
shipping box and 
pallet carton%
}.
In order to increase the diversity and quantity of the image data, we translated the English language search queries to German and Chinese.
The parcel search was performed in English and German for the search engines Google, Yahoo and Bing.
Chinese was used for Baidu.
In total, we collected $21,862$ images of the object of interest.

\paragraph*{Distractor objects}
Since the distractors can be arbitrary objects, we randomly sampled 100 category names from the \shapenetsem{} dataset \cite{savvaSemanticallyenriched3DModels2015} and used these as search query.
This gives us a wide range of object categories, while simultaneously preventing the introduction of a strong bias towards certain categories.
Since it is easier to find suitable distractors, we only performed the search in English and German and focused on Google Image search.
In total we downloaded more than 12,000 images for distractors.

\paragraph*{Background Images}
We did not scrape background images, but instead used images from the SUN397 database \cite{xiaoSUNDatabaseExploring2016}.
We excluded the category \textit{archive} since parcels might be contained in the background, and sampled the scene categories randomly otherwise.
Our training, test, and validation split is done across categories, not image instances, in order to prevent a leakage of background image information.

\subsection{Image Selection}
\label{sec:data:selection}
Scraping images by merely a textual input can yield high quantities of data, however, it is difficult to assess the suitability of each of the images for the dedicated use-case.
Desirable are images, where

\begin{itemize}
    \item the image quality is sufficiently high to enable high-quality image compositions,
    \item the image is a photograph of a real scene,
    \item the background is homogeneous and easy to remove, and
    \item the object of interest is the only object and not occluded.
\end{itemize}

To select such images, we started off by removing all tiny images, i.e. smaller than \SI{80}{kb} in size.
This threshold was determined empirically, trying to prevent the usage of low quality images.
The next step is to analyze the backgrounds of the images.
Since we want to cut out objects automatically in the next step, we discard images with inhomogeneous backgrounds.
This is achieved by analyzing the color variability of the outer frame of the images.
More precisely, we compute the variance of all pixels within a \SI{2}{\%} outer margin of the image, and keep all images where the mean of the variance of the three color channels is smaller than $50$.
Subsequently, we apply the automated background removal tool \textit{Rembg}\footnote{See \href{https://github.com/danielgatis/rembg}{github.com/danielgatis/rembg}} that converts images into masked RGB-A images.
The tool is based on \textit{U$^2$-Net} \cite{qinU2NetGoingDeeper2020} and is used to segment the objects of interest.
Automated background removal is a challenging task, since the pre-processing only removes images with a strong background variance, however, the images might still contain a cluttered scene with multiple objects.
We noticed that especially for difficult images the resulting segmentation masks contain large zones with a high transparency.
This leads to the foreground object smoothly transitioning into the background.
Since this is not desired, we filter out such examples by computing a mask transparency score.
The score is calculated as the percentage of non-zero pixels that have an opacity value below a certain threshold.
This threshold is chosen to be an opacity of \SI{95}{\%}, and we keep images with a mask transparency score smaller than \SI{0.1}{}.
These pre-processing steps are applied equally for all images we collected, \idest objects of interest and distractor objects.
In the case of the object of interest, this reduces our set of images from over $21,000$ to $2,859$.

In order to analyze the impact of the image selection on the quality of the resulting dataset, we create three different datasets from the $2,859$ parcel image candidates. 

\paragraph*{\dsplain{}} This is the base dataset, where in addition to the above-mentioned pre-processing a selection based on mask convexity is introduced.
We compute a mask convexity score, and discard all images with a mask convexity score smaller than \SI{95}{\%}.
We compute the convexity score as the quotient of the area of the biggest contour divided by its convex hull.
It has a total size of $1,321$ instances of parcels.

\paragraph*{\dsmask{}} We only use the annotations for the category "box" of the \openimages{} dataset \cite{kuznetsovaOpenImagesDataset2020}%
\footnote{We use the updated dataset \openimages{} V6 from \href{https://storage.googleapis.com/openimages/web/index.html}{https://storage.googleapis.com/openimages/web/index.html}}, in order to train a \maskrcnn{} \cite{heMaskRCNN2017}.
We employ this \maskrcnn{} to detect whether there is exactly one parcel in a scraped image.
The detection score threshold is set to \SI{95}{\%} and we discard any images with more or less than one detected parcel.
In addition, we use the same mask convexity as described for \dsplain{}.
This dataset consists of $1,066$ instances.

\paragraph*{\dsmanual{}} We revised the $2,859$ candidate images manually, to only select the ones we find suitable, \idest photos of a single parcel with a homogeneous background.
The final dataset contains $854$ instances.

\subsection{Image Generation}
\label{sec:data:generation}
For the generation of the final datasets, we always use the full set of distractor objects and the respective set of parcel objects.
To ensure fair comparability we used the same configuration for all datasets:
We sample between $1$-$4$ objects of interests and $2$-$4$ distractor objects.
We paste these objects at a random position onto the background while maintaining the original size of the background and introducing 2D rotations and scaling of the objects.
In order to guarantee a suitable size of the objects, we limit their scale such that their (relative to the background image) longer side occupies between \SI{15}{\%} and \SI{40}{\%} of the image.
In addition, we allow a maximum upscaling by \SI{20}{\%}, since otherwise the objects potentially become overly blurry.
We set a maximum \gls{iou} of $0.5$ between objects, and reattempt randomly pasting the objects onto the background, if this threshold is crossed.
When a suitable arrangement of objects and distractors is found, we generate four different versions of the same image.
This means, we leave the background, the objects and their positions the same, and only adjust the composition method.
The following blending methods are used: no blending, gaussian blending, motion blur and Poisson blending \cite{perezPoissonImageEditing2003}.
Compared to \textcite{dwibediCutPasteLearn2017}, we add motion blur.
We generate $2,000$ training image configurations and $500$ for each, validation and testing.
Note that the number of images is four times that much, since we generate one image per blending method.

\subsection{Evaluation Dataset: \parcelds{} Real}
\label{sec:data:real}

In order to evaluate the usability of our approach in real world applications, we collected a dataset of parcel photos in various environments.
Our validation dataset comprises $96$ and the test dataset $297$ images.
We describe the data acquisition and the automated annotation process in the following.
Note, that while our focus is on a dataset for instance segmentation, we decided to use an automated approach for the dataset generation, which inherently yields 3D annotations as well.

\subsubsection*{Data acquisition}
\label{sec:data:acquisition}

We built a custom camera rig on which we mounted a Basler Blaze time-of-flight camera and a Stereolab Zed2 stereo camera. The sensor of the Blaze and the center of the Zed2 are aligned vertically.
To allow the transferral of annotations from the depth image of the Blaze to the color images of the Zed2, we calibrated the Blaze with each, the left and the right camera of the Zed2.
For the acquisition of the photos, we mounted the camera rig onto a tripod.
For each image, we additionally collected the background ID and the IDs of the parcels present in the image.
See \reffigure{fig:dataset:real} for exemplary images.

\begin{figure}[h!] 
	\centering
		\includegraphics[width=0.45\linewidth]{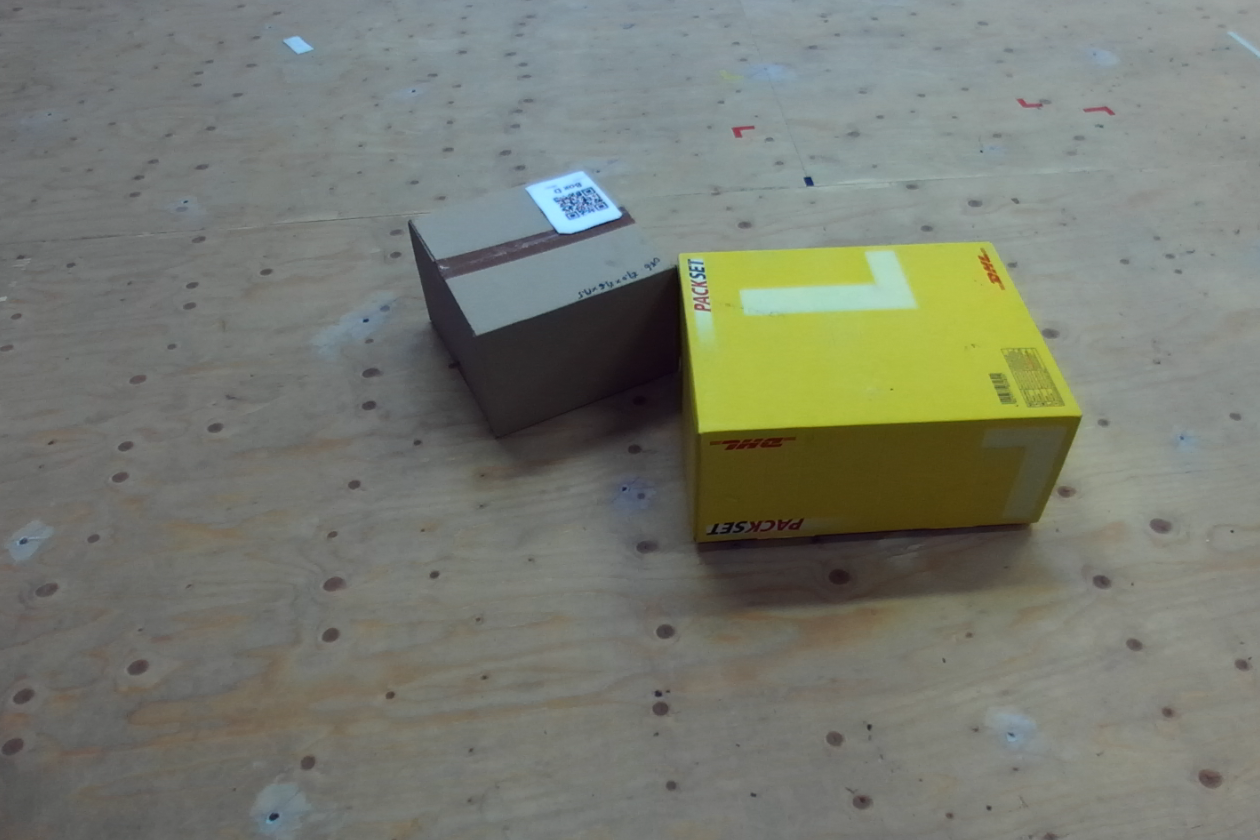} 
		\includegraphics[width=0.45\linewidth]{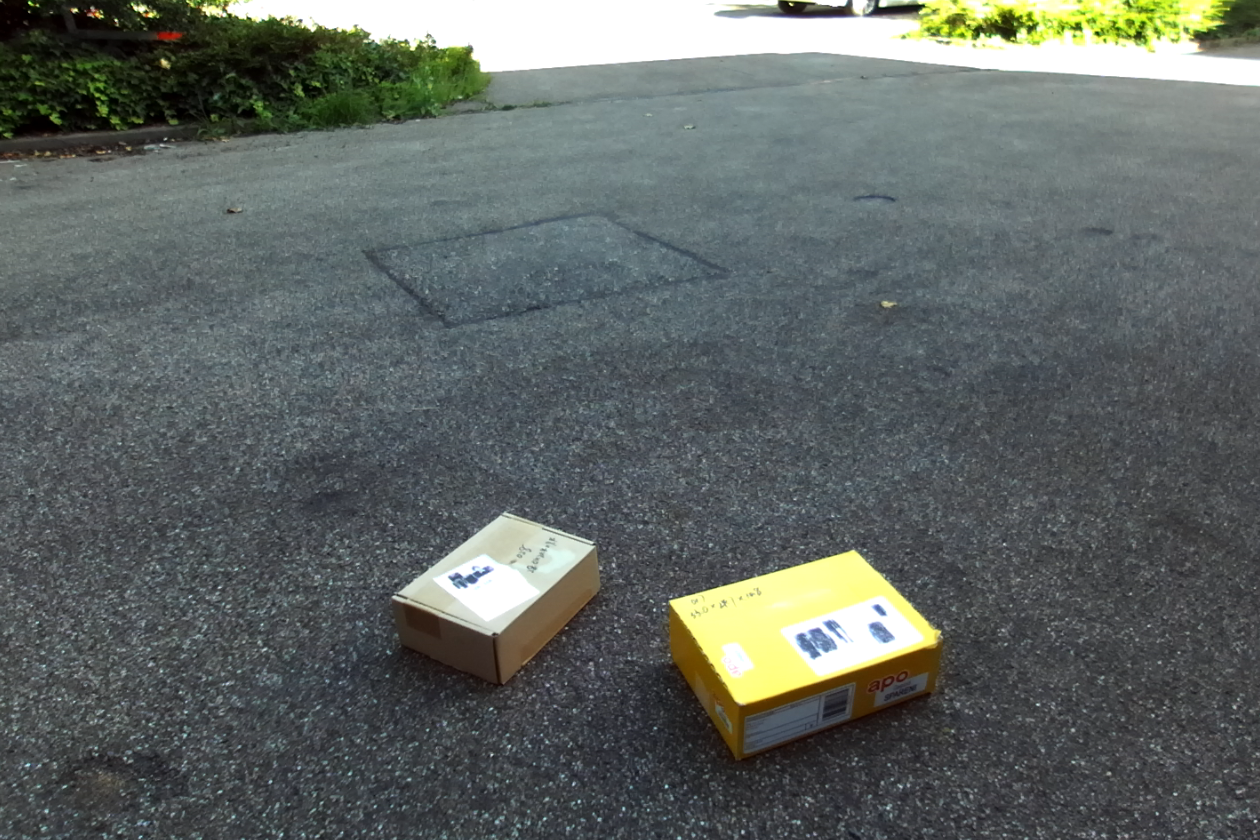}
	\caption{
		Exemplary images of the \dsreal{} dataset.
	}
	\label{fig:dataset:real} 
\end{figure}

\subsubsection*{Annotation generation}
\label{sec:data:annotation}
Starting from the captured RGB-D image as seen in \reffigure{fig:dataset:annnotation:a}, we first applied a plane segmentation approach \cite{furrerIncrementalObjectDatabase2018,grinvaldVolumetricInstanceAwareSemantic2019} (see \reffigure{fig:dataset:annnotation:b}).
To identify the ground plane, we assumed that it is close to the camera and, in comparison with other planes, relatively large.
Using the ground plane, we searched all candidates for parcel top planes by computing the angle between the corresponding normal vectors.
To reliably identify parcels, we discard top plane candidates based on the ground truth parcel dimensions. 
By projecting the remaining parcel top planes onto the ground plane, and fitting the best 3D bounding box with ground truth dimensions around the points using a RANSAC approach \cite{fischlerRandomSampleConsensus1987}, we identify the final parcel annotation as exemplary shown in \reffigure{fig:dataset:annnotation:c}.
These annotations can then be projected onto the color images using the calibration information and we obtain the annotated RGB images as in \reffigure{fig:dataset:annnotation:d}.
We manually revised the dataset to not include erroneous detections.

\begin{figure}[h!] 
	\centering
    \begin{subfigure}[]{0.45\linewidth}
		\centering
		\includegraphics[width=\linewidth]{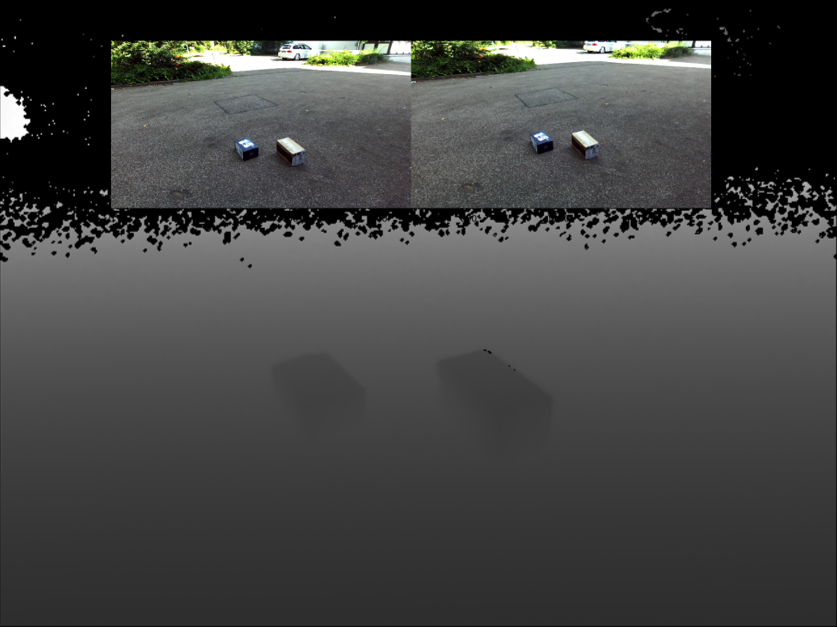} 
		\caption{Raw Input} 
	    \label{fig:dataset:annnotation:a} 
	\end{subfigure}
    \begin{subfigure}[]{0.45\linewidth}
		\centering
		\includegraphics[width=\linewidth]{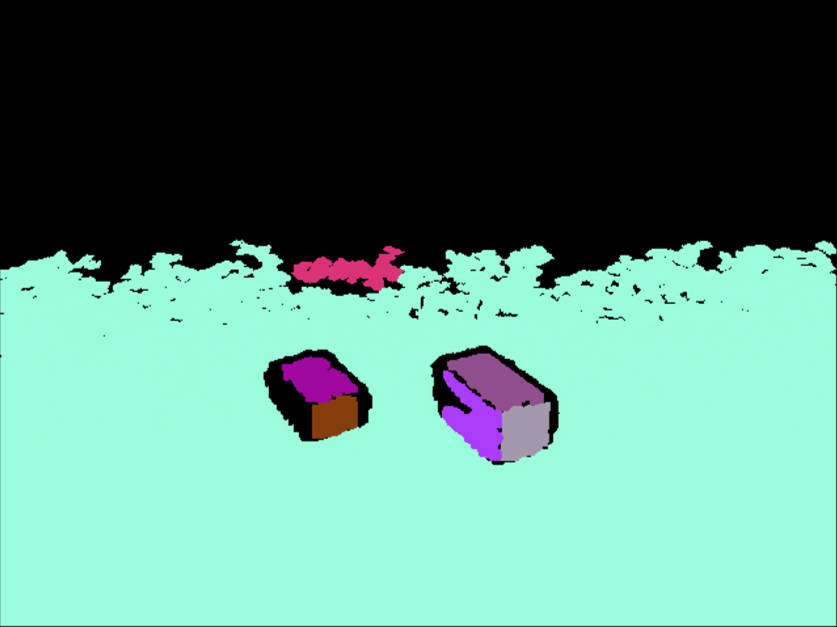} 
		\caption{Plane Segmentation} 
	    \label{fig:dataset:annnotation:b} 
	\end{subfigure}\\
    \vspace{2mm}
    \begin{subfigure}[]{0.45\linewidth}
		\centering
		\includegraphics[width=\linewidth]{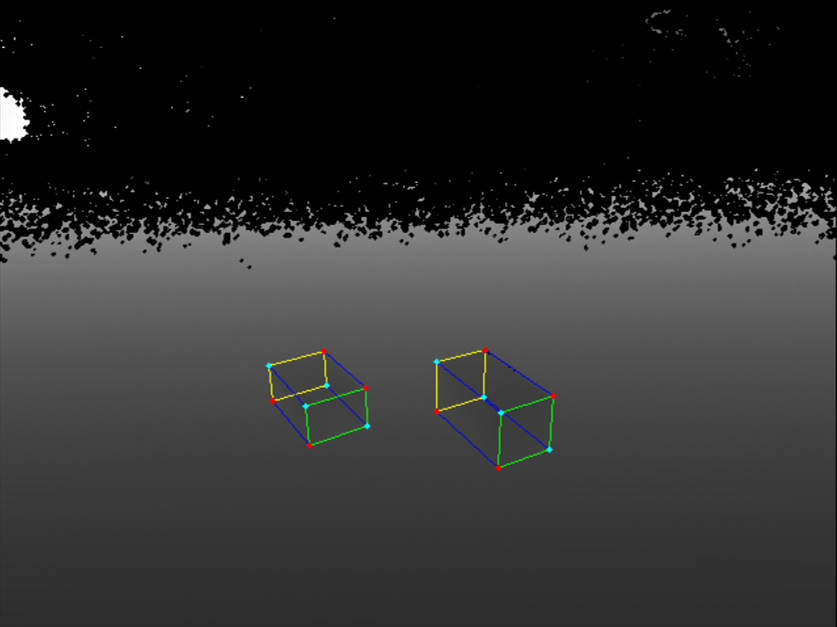} 
		\caption{Annotated RGB-D image} 
	    \label{fig:dataset:annnotation:c} 
	\end{subfigure}
    \begin{subfigure}[]{0.45\linewidth}
		\centering
		\includegraphics[width=\linewidth]{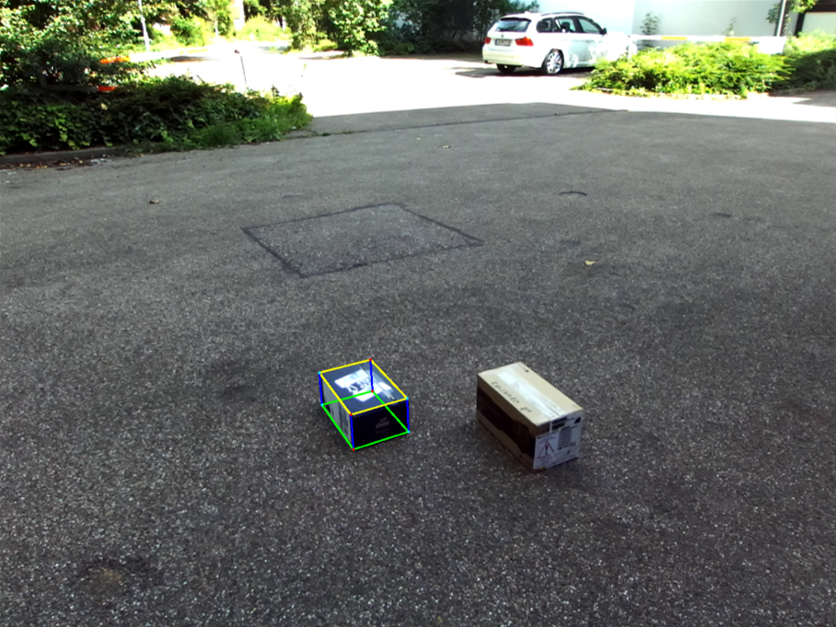} 
		\caption{Annotated RGB image} 
	    \label{fig:dataset:annnotation:d} 
	\end{subfigure}
	\caption{
		Visualization of the annotation generation process:
        (a) Raw input data from the depth and stereo camera,
        (b) plane segmentation result for the RGB-D image,
        (c) resulting annotation on the depth image, and
        (d) annotation that has been transferred onto the RGB image (for one image and one parcel only).
	}
	\label{fig:dataset:annnotation} 
\end{figure}
\section{Evaluation}
\label{sec:eval}

\subsection{Model Configuration}
For all our experiments, we employ a ResNet-50-FPN \cite{linFeaturePyramidNetworks2017} that was pre-trained on the Microsoft COCO dataset \cite{linMicrosoftCOCOCommon2014} as backbone.
The same augmentation techniques are used during training for all datasets.
For all our experiments we use \gls{sgdm} with a batch size of \SI{16}{} and a frozen backbone.
The learning rate schedule is a cosine learning rate schedule \cite{loshchilovSGDRStochasticGradient2017} with an initial learning rate of \SI{0.01}{} and a final learning rate of \SI{0}{} after \SI{15,000}{} iterations. 
Additionally, we apply a linear warm up during the first \SI{1,000}{} iterations.
We select the best model based on the highest segmentation \gls{ap} on the validation set in all our experiments.

\subsection{Comparison of Image Selection Strategies}

\begin{table*}[!htb]
    \centering
    \begin{tabular}{lrrrrrr}
    \toprule
                            &  Box AP &  Box \apfifty{} &  Box \apseventy{} &  Mask AP &  Mask \apfifty{} &  Mask \apseventy{} \\
    \midrule
                \dsplain{} &    \textbf{69.0} &      \textbf{99.0} &      \textbf{86.2} &     \textbf{86.2} &       \textbf{99.0} &       97.4 \\
            \dsmask{} &    63.3 &      98.0 &      80.3 &     84.8 &       98.0 &       \textbf{97.7} \\
                \dsmanual{} &    65.5 &      95.0 &      82.1 &     81.5 &       94.9 &       93.4 \\
    \midrule
                    \openimages{} \cite{kuznetsovaOpenImagesDataset2020} &    \textbf{78.0} &      \textbf{99.0} &      \textbf{97.9} &     85.7 &       \textbf{99.0} &       96.9 \\

    \bottomrule
    \end{tabular}
    \caption{
        Evaluation results on \dsreal{}.
    }
    \label{table:results}
\vspace{-0.5cm}
\end{table*}

We analyze the influence of the three presented image selection methods, by training a \backbone{} on each of the created datasets, and subsequently evaluating their performance on \dsreal{} that was presented in \refsection{sec:data:real}.
Furthermore, we add a baseline to cover the special case when a domain-specific dataset is available.
For our baseline numbers, we train on real photographs of boxes taken from the \openimages{} dataset category "box".
This training dataset contains $2,086$ instances.
Note, that the \openimages{} definition of box is broader than the one used for our manual image selection.
The results are summarized in \reftable{table:results}.
We see that all three methods for image selection, \idest no image selection (Plain), image selection by \maskrcnn{} and manual image selection can be used to generate suitable datasets.
All resulting datasets allow a transfer from synthetic to real data as indicated by the Box \apseventy{}, which is above \SI{80}{} in all cases.
For the case of object detection, however, training on the relevant subset of the \openimages{} dataset yields the best results as implicated by the Box \apseventy{}.
While for the Box \apfifty{}, results are comparable across the different datasets, \openimages{} clearly outperforms the other datasets on Box \apseventy{}.
This might be due to the broader and thus, more diverse definition of the category of interest, box.
The same argument can be applied to the comparison of the three image selection methods: the Plain variant performs best for object detection and segmentation and at the same time has the broadest definition of the category box, since no object-specific filtering is applied.
The fact that the image domain of the training data should contain the one of the test data to get highest performance during transfer learning, was analyzed by \textcite{mensinkFactorsInfluenceTransfer2021} and can be confirmed for our application.

The results for the task of image segmentation are different.
All datasets have a Mask \apseventy{} above \SI{90}{} and thus, perform very well on the test dataset.
Differences between training on the Plain dataset, the CNN dataset and \openimages{} are rather small, only the manually selected dataset performs worse.
The best dataset according to the Mask AP is the Plain dataset.
We cannot generalize these findings to arbitrary tasks, however, it is noteworthy that contrary to human intuition, a cautious cherry-picking of instance examples does not always yield the best performance.
Finally, we trained the \backbone{} on both \openimages{} and \parcelds{} Plain combined.
The results on the real test dataset are a Box AP of \SI{72.6}{} and a Mask AP of \SI{86.5}{}.
Thus, training on the combination of the two datasets is beneficial considering Mask AP.

\subsection{Ablation Study}

Since \textcite{dwibediCutPasteLearn2017} and \textcite{ghiasiSimpleCopyPasteStrong2021} do not agree on the importance of blending methods, we performed an ablation study to check which finding holds true in our use-case.
\citeauthor{ghiasiSimpleCopyPasteStrong2021} question the importance of using blending methods, whereas \citeauthor{dwibediCutPasteLearn2017} claim, that blending methods are important for the quality of the dataset.
We obtain a Box AP of \SI{51.2}{} and a Mask AP of \SI{70.9}{}, when training on \parcelds{} Manual, without using any blending methods.
Since this is a considerable drop in performance, compared to the case with blending methods, we argue that blending methods are an important factor.
Further, the effect that \citeauthor{ghiasiSimpleCopyPasteStrong2021} observed, probably stems from the fact that they augment existing annotated images.
These images inherently contain annotated objects where no local pasting artifacts are present in addition to pasted objects with local artifacts and thus, the model cannot focus on pasting artifacts as main visual cues.

\section{Conclusion}
\label{sec:conclusion}

We presented a fully automated and flexible pipeline for the generation of domain-specific artificial instance segmentation datasets.
Our pipeline consists of four steps: 
(1) collecting images by web scraping, 
(2) image selection with three different techniques, 
(3) image arrangement by randomly scaling and placing objects of interest and distractors onto a background and finally, 
(4) image blending in order to remove local artifacts from pasting.
These steps follow the ideas of \textcite{dwibediCutPasteLearn2017} with the exception of the image acquisition phase, \idest the first two steps.
We provide the source code to facilitate the dataset generation for other use-cases.

We compared three different image selection strategies and found that the most accurate one, \idest manual image selection does not yield the best performing training dataset.
In fact, for our use-case the \dsplain{} dataset performs best despite requiring only primitive object-agnostic pre-processing.
While the results were convincing for our case study of parcel detection, performance does strongly depend on the quality and quantity of images that can be obtained in the image acquisition step.
Therefore, the image acquisition must be adjusted for specialized objects, as images picturing them cannot be retrieved by search engines in the desired quantity.
For example, a pipeline similar to the one of \textcite{dwibediCutPasteLearn2017} could be used for the objects of interest, while the acquisition of diverse distractor objects can still be performed as presented in this work.
In general, there will usually be a trade-off between the invested time and the quality of your dataset.

Note, that scraping arbitrary images from the internet will often mean that those images are protected by Copyright.
Keep in mind potential infringements of Copyrights when using or providing a dataset.

It would be interesting to further investigate the suitability of our approach for different use-cases to verify if and to what extent our results generalize.
In addition to that, the influence of the different blending methods and especially a systematic analysis of method combinations would be interesting.

~\\
{
\small
\textbf{Acknowledgement: }
This work was supported by the German Federal Ministry of Education and Research (BMBF) as part of the project PfleDaKi (16SV8887).
}
\endgroup
\printbibliography

\end{document}